\documentclass[letterpaper, 10 pt, conference]{ieeeconf}

\IEEEoverridecommandlockouts
\usepackage{amsmath}
\usepackage{amssymb}
\usepackage{graphicx}
\usepackage{booktabs}
\usepackage{multirow}
\usepackage{cite}
\usepackage{url}

\title{\LARGE \bf
Coordinating from Memory: Graph-Structured Experience Reuse for Multi-Agent Adaptation in Dynamic Manufacturing
}

\author{Chengxiao Dai$^{1}$, Zhanhui Lin$^{1}$, Zhaokun Yan$^{3}$, Youyang Ni$^{4}$, Chenjun Lei$^{5}$, Luyan Zhang$^{2,*}$%}
\thanks{$^{*}$Corresponding author ({\tt\small zhang.luya@northeastern.edu}).}
\thanks{$^{1}$School of Computer Science, University of Sydney, Sydney, Australia.}%
\thanks{$^{2}$Khoury College of Computer Sciences, Northeastern University, Boston, USA.}%
\thanks{$^{3}$School of Computation, Information and Technology, Technical University of Munich, Munich, Germany.}%
\thanks{$^{4}$School of Life and Environmental Sciences, University of Sydney, Sydney, Australia.}%
\thanks{$^{5}$College of Business and Economics, Australian National University, Canberra, Australia.}%
}

\begin{document}

\maketitle
\thispagestyle{empty}
\pagestyle{empty}

%%%%%%%%%%%%%%%%%%%%%%%%%%%%%%%%%%%%%%%%%%%%%%%%%%%%%%%%%%%%%%%%%%%%%%%%%%%%%%%%
\begin{abstract}

% Dynamic manufacturing environments require multi-agent systems to coordinate effectively under frequent operational disturbances such as machine failures, urgent job arrivals, and processing time fluctuations. Existing multi-agent reinforcement learning approaches treat each disturbance episode independently, discarding valuable coordination experience that could accelerate future adaptation. In this paper, we propose a Graph-Structured Experiential Memory (GSEM) framework for multi-agent coordination in dynamic manufacturing. The framework encodes historical coordination episodes as heterogeneous relational graphs that capture task dependencies, machine states, and inter-agent collaboration patterns. When a new disturbance occurs, a graph neural network-based retrieval mechanism identifies structurally similar past episodes, enabling experience-guided policy adaptation rather than learning from scratch. Experiments on dynamic flexible job-shop scheduling benchmarks with three disturbance types show that GSEM reduces makespan by $4.1$\%--$10.0$\% and adaptation time by $33$\%--$38$\% compared to state-of-the-art baselines, with the advantage increasing under higher disturbance frequency. Ablation studies and cross-disturbance transfer experiments further confirm the contribution of graph-structured encoding, memory retrieval, and the generalizability of learned coordination patterns.

Dynamic manufacturing environments require multi-agent systems to coordinate effectively under frequent operational disturbances such as machine failures, urgent job arrivals, and processing time variations. Existing multi-agent reinforcement learning approaches treat each disturbance episode independently, discarding valuable coordination experience that could accelerate future adaptation. In this paper, we propose a Graph-Structured Experiential Memory (GSEM) framework for multi-agent coordination in dynamic manufacturing. The framework encodes historical coordination episodes as heterogeneous relational graphs that capture task dependencies, machine states, and inter-agent collaboration patterns. When a new disturbance occurs, a graph neural network-based retrieval mechanism identifies structurally similar past episodes, enabling experience-guided policy adaptation rather than learning from scratch. Experiments on dynamic flexible job-shop scheduling benchmarks with three disturbance types show that GSEM reduces makespan by $4.1$\%--$10.0$\% and adaptation time by $33$\%--$38$\% compared to the strongest memory-augmented baseline, with the advantage increasing under higher disturbance frequency. Ablation studies and cross-disturbance transfer experiments further validate the necessity of graph-structured encoding and similarity-based retrieval, and demonstrate the cross-disturbance generalizability of learned coordination patterns.
\end{abstract}

%%%%%%%%%%%%%%%%%%%%%%%%%%%%%%%%%%%%%%%%%%%%%%%%%%%%%%%%%%%%%%%%%%%%%%%%%%%%%%%%
\section{INTRODUCTION}

Modern manufacturing systems in the Industry 4.0 era are characterized by high product variety, short production cycles, and frequent uncertainty, making shop-floor scheduling much more challenging than in traditional static settings \cite{c1_serrano2021}. In practice, production plans are often disrupted by machine breakdowns, urgent job arrivals, material shortages, and processing-time fluctuations, which may propagate across machines, jobs, and production stages \cite{c2_ouelhadj2009}. Dynamic scheduling should therefore be viewed not simply as repeated local rescheduling, but as an online coordination problem in which distributed decision makers detect disturbances and adjust actions to maintain production performance \cite{c2_ouelhadj2009,c3_leitao2009}.

Multi-agent systems provide a natural foundation for distributed manufacturing control by decomposing decision-making across machines or production units while preserving system-level coordination objectives \cite{c3_leitao2009}. Based on this paradigm, reinforcement learning (RL) and multi-agent reinforcement learning (MARL) have shown strong potential for dynamic scheduling, especially in flexible job-shop environments \cite{c4_zhou2021,c9_zhang2024}. Recent studies have reported promising results using cooperative MARL methods in dynamic manufacturing scenarios \cite{c5_gui2024,c11_rashid2020,c12_yu2022}. However, most existing approaches still treat disturbance episodes as largely independent. When a new disruption occurs, agents usually rely on the current policy or replayed transitions, which mainly improve training efficiency rather than support structured adaptation during execution \cite{c7_lin1992,c15_schaul2016}. As a result, current MARL methods often show slow adaptation, weak reuse of coordination experience, and limited generalization across heterogeneous disturbances \cite{c4_zhou2021,c5_gui2024}.

This limitation is important because disturbances in manufacturing are rarely isolated. A local machine failure may trigger downstream waiting, queue imbalance, and coordination conflicts among multiple agents \cite{c2_ouelhadj2009}. Effective adaptation depends on whether the system can recall and reuse coordination patterns from similar past situations, much as human operators handle disruptions by recalling similar incidents \cite{c6_nonaka1995}. A key challenge is representation: manufacturing coordination is inherently relational, involving precedence constraints, resource competition, and inter-agent dependencies that are difficult to preserve using flat state-action vectors \cite{c7_lin1992,c15_schaul2016}. Graph neural networks (GNNs) can model such relational structures effectively \cite{c18_zhang2020,c19_song2023,c20_park2021,c21_wan2024hgnn,c22_jiang2020}, but have mainly been used for current-state encoding rather than experience storage.

This paper proposes a \textit{Graph-Structured Experiential Memory} (GSEM) framework that stores coordination episodes as heterogeneous relational graphs and retrieves structurally similar past experiences for rapid adaptation. The key insight is that what needs to be reused is not a raw past state, but a \textit{reusable recovery motif}: the coordinated rerouting pattern after a breakdown, the queue rebalancing after an urgent job arrival, or the conflict resolution under processing time variation. These motifs are inherently relational, and flat episodic memory cannot preserve them.

The main contributions of this paper are:
\begin{itemize}
    \item We propose graph-structured coordination memory for disturbance adaptation in multi-agent manufacturing, where past recovery episodes are stored as heterogeneous relational graphs that preserve the structural dependencies essential for experience reuse.
    \item We design a structure-aware memory retrieval mechanism that identifies relevant past episodes through GNN-based graph similarity with quality-recency weighting, rather than flat nearest-neighbor matching.
    \item We integrate retrieved experience into multi-agent coordination through gated attention fusion and memory-conditioned communication, enabling agents to collectively exploit past recovery motifs for faster adaptation.
\end{itemize}

%%%%%%%%%%%%%%%%%%%%%%%%%%%%%%%%%%%%%%%%%%%%%%%%%%%%%%%%%%%%%%%%%%%%%%%%%%%%%%%%
\section{RELATED WORK}

\subsection{Multi-Agent Coordination in Manufacturing}

Multi-agent systems have been widely used in manufacturing scheduling and control because of their modularity, scalability, and fault tolerance \cite{c3_leitao2009}. Early studies mainly relied on rule-based coordination or contract-net-style negotiation, in which agents allocated tasks through predefined interaction protocols \cite{c8_smith1980}. Although effective in stable environments, these approaches are less adaptable to disruptions such as machine failures, urgent job arrivals, and other dynamic events.

With the development of learning-based scheduling, RL has been increasingly applied to manufacturing decision-making. Single-agent RL methods typically formulate scheduling as a Markov decision process and learn a centralized dispatching policy for the shop floor \cite{c9_zhang2024}. However, they often face scalability challenges as the numbers of jobs, machines, and decisions increase. MARL distributes decision-making across multiple machines or production units \cite{c4_zhou2021}. Representative approaches include independent learning \cite{c10_tan1993}, value decomposition methods such as QMIX \cite{c11_rashid2020}, and cooperative policy optimization methods such as Multi-Agent Proximal Policy Optimization (MAPPO) \cite{c12_yu2022}. Recent studies have also shown the potential of adaptive multi-agent scheduling in dynamic manufacturing environments \cite{c5_gui2024}.

Most MARL-based scheduling methods focus on reacting to the current state rather than reusing coordination experience across disturbances. When disruptions occur, agents usually rely on learned policies or replay mechanisms, which may limit adaptation and generalization.

\subsection{Memory-Augmented Learning}

Memory augmentation has been widely studied to improve the adaptability of neural decision systems. Early architectures such as Neural Turing Machines and Differentiable Neural Computers introduced differentiable read-write mechanisms that allow models to store and retrieve information beyond fixed network parameters \cite{c13_graves2014,c14_graves2016}. In reinforcement learning, experience replay and prioritized replay are standard techniques for improving sample efficiency by reusing past transitions during training \cite{c7_lin1992,c15_schaul2016}.

More recently, episodic memory mechanisms have been proposed to support faster adaptation by storing and retrieving past episodes rather than isolated transitions \cite{c16_blundell2016,c17_pritzel2017}. These methods typically retrieve historical experiences through nearest-neighbor matching in an embedding space, allowing agents to recall useful behaviors in similar situations.

However, these methods store experience as fixed-dimensional vectors or flat transition sequences, which cannot preserve the relational dependencies among jobs, operations, machines, and agents that are intrinsic to manufacturing coordination.

\subsection{Graph Neural Networks in Manufacturing}

Graph neural networks have emerged as an effective tool for modeling relational structures in manufacturing and scheduling problems. Job-shop and flexible job-shop scheduling naturally involve precedence constraints, machine-operation compatibility, and resource conflicts, making graph-based representations suitable for learning scheduling policies \cite{c18_zhang2020}. Several studies have used GNNs to encode scheduling states and learn dispatching decisions, showing strong performance and generalization across problem instances \cite{c19_song2023,c20_park2021}. Heterogeneous graph neural networks further improve representational expressiveness by distinguishing different node and edge types, such as jobs, machines, operations, and resource relations \cite{c21_wan2024hgnn}.

In multi-agent settings, graph-based learning has also been used to model communication and coordination structures among agents \cite{c22_jiang2020}. By propagating information along relational edges, GNNs allow agents to reason about neighboring states and interactions more effectively than flat feature encodings.

Although graph-based methods capture relational structure effectively, they are mainly used to encode the current system state for reactive decision-making; their use as memory units for storing and retrieving historical coordination episodes has received much less attention.

%%%%%%%%%%%%%%%%%%%%%%%%%%%%%%%%%%%%%%%%%%%%%%%%%%%%%%%%%%%%%%%%%%%%%%%%%%%%%%%%
\section{PROPOSED METHOD}

\begin{figure*}[t]
    \centering
    \includegraphics[width=\textwidth]{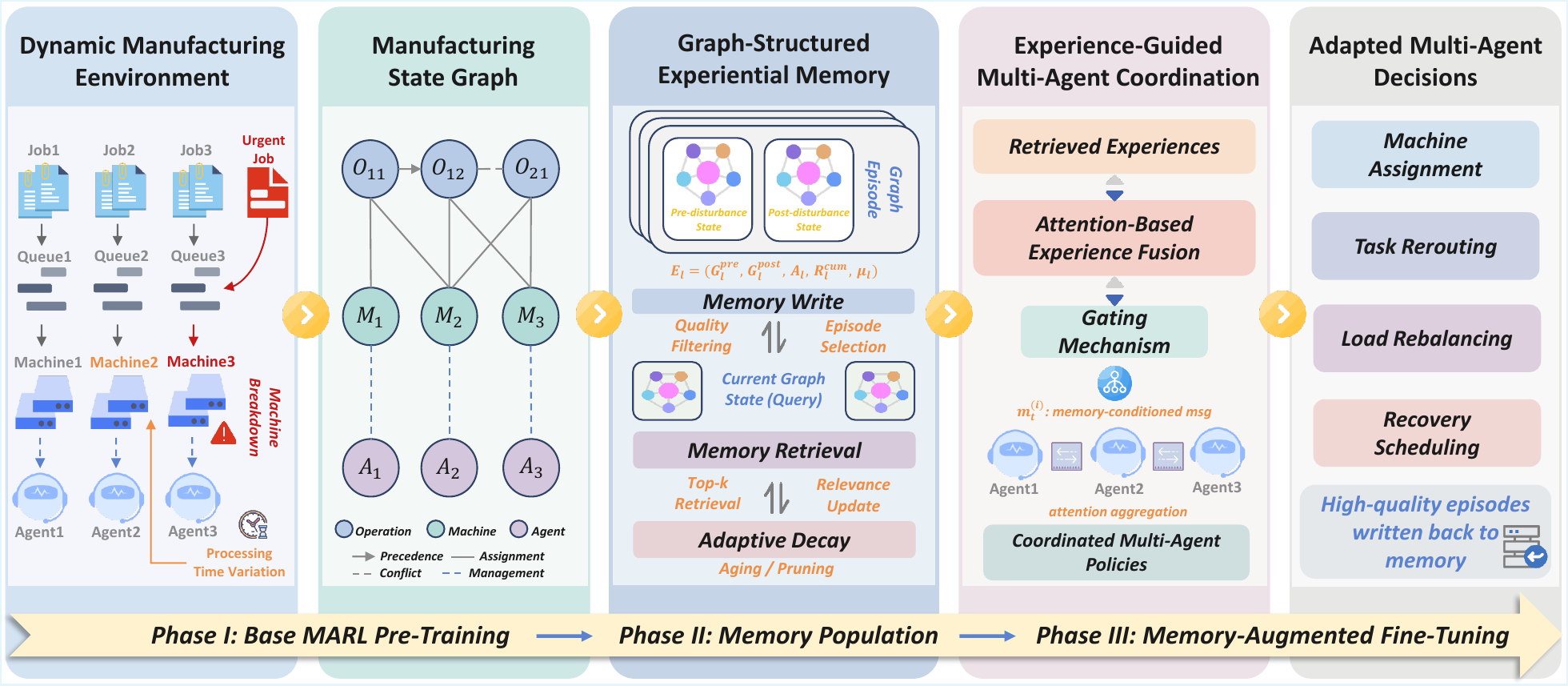}
    \caption{Overall framework of the proposed GSEM system.}
    \label{fig:framework}
\end{figure*}

The goal of GSEM is not only to improve current-state encoding, but to store and retrieve reusable coordination structures from past disturbance recovery episodes. We consider a Dynamic Flexible Job-Shop Scheduling Problem (DFJSP) with $n$ jobs $\mathcal{J}$ and $m$ machines $\mathcal{M}$, where each job consists of ordered operations with machine-dependent processing times. Dynamic disturbances, including machine breakdowns, urgent job arrivals, and processing time variations, occur stochastically during execution. We model this as a Decentralized Partially Observable Markov Decision Process (Dec-POMDP) with $N$ scheduling agents, each managing a disjoint subset of machines. The goal is to find a joint policy $\boldsymbol{\pi}^*$ that minimizes the expected makespan $C_{\max}$ under disturbances. At each step $t$, each agent receives a shared reward composed of a progress term and a completion penalty:
\begin{equation}
r_t = -\alpha \cdot \Delta C_{\max}(t) - \beta \cdot \sum_{j \in \mathcal{J}} \max(0,\, C_j - d_j)
\end{equation}
where $\Delta C_{\max}(t)$ is the incremental makespan change, $C_j$ is the completion time of job $j$, $d_j$ is its due date, and $\alpha, \beta$ are weighting coefficients.

\subsection{Manufacturing State Graph}

We represent the manufacturing state at each decision point as a heterogeneous graph $\mathcal{G}_t = (\mathcal{V}_t, \mathcal{E}_t)$ with three node types (operation, machine, and agent) and four edge types: \textit{precedence} (operation sequencing within jobs), \textit{assignment} (operation-to-compatible-machine links weighted by processing time), \textit{management} (machine-to-agent ownership), and \textit{conflict} (operations competing for the same machine). A Relational Graph Convolutional Network (R-GCN) encoder produces node embeddings through $L$ message-passing layers, and a \textbf{type-aware readout} generates a graph-level embedding by mean-pooling each node type separately and concatenating:
\begin{equation}
\mathbf{z}_{\mathcal{G}} = \left[\bar{\mathbf{h}}^{\text{op}} \,\big\|\, \bar{\mathbf{h}}^{\text{mc}} \,\big\|\, \bar{\mathbf{h}}^{\text{ag}}\right], \quad \bar{\mathbf{h}}^{\tau} = \frac{1}{|\mathcal{V}^{\tau}|} \sum_{v \in \mathcal{V}^{\tau}} \mathbf{h}_v^{(L)}
\end{equation}
This preserves the distinct semantics of different node types that would be lost in naive global pooling.

\subsection{Graph-Structured Experiential Memory}

The memory pool $\mathcal{P}$ stores bounded past coordination episodes. Each entry is a tuple $E_l = (\mathcal{G}_l^{\text{pre}}, \mathcal{G}_l^{\text{post}}, \mathbf{A}_l, R_l^{\text{cum}}, \boldsymbol{\mu}_l)$ containing the state graphs before and after a disturbance, the joint action sequence during recovery, the cumulative reward, and metadata (disturbance type, severity, timestamp).

\textbf{Memory Write.}
Episodes are filtered by two criteria before storage. First, episode boundaries are identified by detecting disturbance onset (a $|\Delta C_{\max}|$ spike beyond $2\sigma$ of recent history) and recovery completion (per-step reward returning within $1\sigma$ of its pre-disturbance moving average). Second, only episodes whose cumulative reward satisfies $R_l^{\text{cum}} \geq \bar{R} - \kappa \cdot \sigma_R$ are retained, where $\bar{R}$ and $\sigma_R$ are running statistics and $\kappa$ controls selectivity.

\textbf{Memory Retrieval.}
When a disturbance occurs, the current state graph $\mathcal{G}_t$ is encoded and compared against stored entries using cosine similarity with a quality-recency weight:
\begin{equation}
\text{sim}(\mathcal{G}_t, E_l) = \frac{\mathbf{z}_{\mathcal{G}_t}^{\top} \mathbf{z}_{\mathcal{G}_l^{\text{pre}}}}{\|\mathbf{z}_{\mathcal{G}_t}\| \cdot \|\mathbf{z}_{\mathcal{G}_l^{\text{pre}}}\|} \cdot \tilde{R}_l \cdot e^{-\eta(t - \tau_l)}
\end{equation}
where $\tilde{R}_l = (R_l^{\text{cum}} - R_{\min}) / (R_{\max} - R_{\min}) \in [0,1]$ is a min-max-normalized quality score (needed because cumulative rewards are non-positive under Eq.~(1)), and $R_{\min}, R_{\max}$ are running statistics over $\mathcal{P}$. Top-$K$ episodes are then selected via maximal marginal relevance (MMR) to balance relevance and diversity.

\textbf{Adaptive Decay.}
Each entry maintains a relevance score $\rho_l$ updated as:
\begin{equation}
\rho_l \leftarrow \rho_l \cdot \exp\!\left(-\eta_d (t - t_l^{\text{last}})\right) + \mathbb{1}[E_l \in \mathcal{S}_t] \cdot \Delta\rho
\end{equation}
where $\mathcal{S}_t$ is the top-$K$ retrieved set at step $t$: retrieved entries receive a bonus $\Delta\rho$ and unused entries decay over time. When $|\mathcal{P}|$ exceeds capacity, the lowest-$\rho$ entry is evicted.

\subsection{Experience-Guided Coordination}

Retrieved episodes are integrated into each agent's policy through attention-based fusion and gating. For each retrieved episode $E_l$, the system exploits all stored components: the pre-disturbance graph $\mathcal{G}_l^{\text{pre}}$ provides situational context, the post-recovery graph $\mathcal{G}_l^{\text{post}}$ encodes the target coordination state, and the action sequence $\mathbf{A}_l$ supplies the recovery strategy. These are fused into a unified episode representation:
\begin{equation}
\mathbf{e}_l = \mathbf{W}_e \left[ \mathbf{z}_{\mathcal{G}_l^{\text{pre}}} \| \mathbf{z}_{\mathcal{G}_l^{\text{post}}} \| f_a(\mathbf{A}_l) \right] + \mathbf{b}_e
\end{equation}
where $f_a(\cdot)$ is a temporal mean-pooling over the action sequence embeddings.

\textbf{Attention Fusion.}
Each agent $i$'s R-GCN embedding $\mathbf{h}_{v_i}^{(L)}$ serves as query, while the episode representations $\{\mathbf{e}_l\}_{l=1}^{K}$ serve as keys and values via cross-attention with learned projection matrices $\mathbf{W}_Q, \mathbf{W}_K, \mathbf{W}_V \in \mathbb{R}^{d \times d}$:
\begin{equation}
\begin{aligned}
\alpha_l^{(i)} &= \text{softmax}\!\left(\frac{(\mathbf{W}_Q \mathbf{h}_{v_i}^{(L)})^\top (\mathbf{W}_K \mathbf{e}_l)}{\sqrt{d}}\right), \\
\mathbf{c}_t^{(i)} &= \sum_{l=1}^{K} \alpha_l^{(i)} \mathbf{W}_V \mathbf{e}_l
\end{aligned}
\end{equation}

\textbf{Gating Mechanism.}
A learned gate $\mathbf{g}_t^{(i)}$ adaptively balances between current state and retrieved experience:
\begin{equation}
\mathbf{g}_t^{(i)} = \sigma\!\left(\mathbf{W}_g [\mathbf{h}_{v_i}^{(L)} \| \mathbf{c}_t^{(i)}] + \mathbf{b}_g\right)
\end{equation}
\begin{equation}
\tilde{\mathbf{h}}_t^{(i)} = \mathbf{g}_t^{(i)} \odot \mathbf{c}_t^{(i)} + (1 - \mathbf{g}_t^{(i)}) \odot \mathbf{h}_{v_i}^{(L)}
\end{equation}
The gate tends toward experience reliance immediately after a disturbance and shifts toward the current policy as the agent adapts.

\textbf{Memory-Conditioned Communication.}
To enable collective experience-driven coordination, each agent $i$ broadcasts a memory-conditioned message $\mathbf{m}_t^{(i)} = \mathbf{W}_m \tilde{\mathbf{h}}_t^{(i)}$ and aggregates incoming messages via attention over its neighbor set $\mathcal{N}(i)$, defined as agents whose machines share at least one operation with those of $i$ via \textit{assignment} or \textit{conflict} edges in $\mathcal{G}_t$:
\begin{equation}
\begin{aligned}
\beta_{ij} &= \text{softmax}\!\left(\frac{(\tilde{\mathbf{h}}_t^{(i)})^\top \mathbf{m}_t^{(j)}}{\sqrt{d}}\right) \\
\bar{\mathbf{m}}_t^{(i)} &= \sum_{j \in \mathcal{N}(i)} \beta_{ij} \mathbf{m}_t^{(j)}
\end{aligned}
\end{equation}
The final policy $\pi_i(a_t^{(i)} | \cdot)$ takes the concatenation $[\tilde{\mathbf{h}}_t^{(i)} \| \bar{\mathbf{m}}_t^{(i)}]$ as input.

\subsection{Training Procedure}

Training proceeds in three phases: (I)~base policy pre-training with MAPPO \cite{c12_yu2022} without memory; (II)~memory population by rolling out the Phase~I policy, with $\bar{R}, \sigma_R$ in the quality filter initialized from the Phase~I reward distribution; (III)~end-to-end fine-tuning with memory retrieval active and $\bar{R}, \sigma_R$ continuously updated. In Phase~III, the R-GCN encoder is additionally trained with a contrastive loss that pulls graphs of the same disturbance type together, where $\mathcal{G}^+$ is sampled within $\pm 5$ recovery steps of $\mathcal{G}_t$ and $\{\mathcal{G}_l\}_{l=1}^{B}$ are negatives from the current mini-batch:
\begin{equation}
L_{\text{cont}} = -\log \frac{\exp(\text{sim}(\mathbf{z}_{\mathcal{G}_t}, \mathbf{z}_{\mathcal{G}^+}) / \tau)}{\sum_{l=1}^{B} \exp(\text{sim}(\mathbf{z}_{\mathcal{G}_t}, \mathbf{z}_{\mathcal{G}_l}) / \tau)}
\end{equation}
During online adaptation, the agent detects disturbances, retrieves top-$K$ episodes via Eqs.~(3)--(4), applies experience fusion via Eqs.~(5)--(9), and writes high-quality recovery episodes back to $\mathcal{P}$. Parameters are updated with $L_{\text{MAPPO}} + c_2 L_{\text{cont}}$.

%%%%%%%%%%%%%%%%%%%%%%%%%%%%%%%%%%%%%%%%%%%%%%%%%%%%%%%%%%%%%%%%%%%%%%%%%%%%%%%%
\section{EXPERIMENTS}

\subsection{Experimental Setup}

We primarily evaluate GSEM on DFJSP instances with $10$ jobs, $10$ machines, and $5$ scheduling agents, a scale consistent with widely used DFJSP benchmarks in the literature \cite{c18_zhang2020,c19_song2023,c20_park2021}. Each job consists of $3$ to $6$ operations, each compatible with $2$ to $4$ machines, with processing times sampled from $\mathcal{U}(1,20)$. Three types of dynamic disturbances are injected during execution: machine breakdowns with failure rate $\lambda{=}0.03$ per step and downtime drawn from $\mathcal{U}(5,20)$, urgent job arrivals at rate $0.05$ per step, and processing time variations with perturbation factor $\delta\sim\mathcal{U}(0.8,1.2)$. Three disturbance frequency levels are tested: Low, Medium, and High, with $2$--$3$, $5$--$7$, and $10$+ events per episode. \textit{Adapt.} counts steps from disturbance onset until per-step reward recovers to within $1\sigma$ of its pre-disturbance moving average.

We compare against shortest processing time (SPT) and earliest due date (EDD) dispatching rules, a genetic algorithm (GA) with reactive rescheduling, centralized proximal policy optimization (PPO), and MARL methods including QMIX \cite{c11_rashid2020}, MAPPO \cite{c12_yu2022}, and MAPPO+ER with flat experience replay. We further include two adapted baselines: GNN-Dispatch, a GNN-based reactive scheduler adapted from \cite{c19_song2023} without memory, and MAPPO+Epi, which augments MAPPO with flat episodic memory following \cite{c17_pritzel2017}. The R-GCN encoder has $L{=}3$ layers with $128$-dimensional hidden features, and the policy network is a $2$-layer multi-layer perceptron (MLP) with $256$ units. We train with Adam at learning rate $3{\times}10^{-4}$, discount factor $\gamma{=}0.99$, and PPO clipping $\epsilon{=}0.2$ for $20$K Phase~I and $30$K Phase~II+III episodes. The memory pool capacity is $|\mathcal{P}|_{\max}{=}500$ with retrieval size $K{=}5$. All results are averaged over $20$ independent seeds on a server with dual AMD EPYC 9654 CPUs, $2$\,TB of DDR5 RAM, and $4{\times}$\,NVIDIA RTX 5090 GPUs. The contrastive loss weight is set to $c_2{=}0.1$, the decay rate to $\eta_d{=}0.01$, and the quality threshold parameter to $\kappa{=}1.0$. The reward coefficients are $\alpha{=}1.0$ and $\beta{=}0.5$.

\begin{table*}[t]
\caption{Performance Comparison on DFJSP ($10\times10$, $5$ Agents). Makespan ($C_{\max}$), Adaptation Speed (Adapt., steps), and Machine Utilization ($U$).}
\label{tab:main}
\centering
\resizebox{\textwidth}{!}{
\begin{tabular}{l cc cc cc cc cc cc}
\toprule
\multirow{4}{*}{Method} & \multicolumn{6}{c}{\textbf{Single Disturbance (Medium Freq.)}} & \multicolumn{6}{c}{\textbf{Mixed Disturbances (All Three Types)}} \\
\cmidrule(lr){2-7} \cmidrule(lr){8-13}
& \multicolumn{2}{c}{Breakdown} & \multicolumn{2}{c}{Urgent Job} & \multicolumn{2}{c}{Time Variation} & \multicolumn{2}{c}{Low Freq.} & \multicolumn{2}{c}{Medium Freq.} & \multicolumn{2}{c}{High Freq.} \\
\cmidrule(lr){2-3} \cmidrule(lr){4-5} \cmidrule(lr){6-7} \cmidrule(lr){8-9} \cmidrule(lr){10-11} \cmidrule(lr){12-13}
& $C_{\max}$ & Adapt. & $C_{\max}$ & Adapt. & $C_{\max}$ & Adapt. & $C_{\max}$ & $U$(\%) & $C_{\max}$ & $U$(\%) & $C_{\max}$ & $U$(\%) \\
\midrule
SPT & 312$\pm$19 & -- & 298$\pm$14 & -- & 267$\pm$13 & -- & 285$\pm$15 & 68.2 & 334$\pm$19 & 61.5 & 398$\pm$27 & 53.8 \\
EDD & 325$\pm$20 & -- & 285$\pm$15 & -- & 274$\pm$12 & -- & 291$\pm$15 & 66.8 & 341$\pm$23 & 59.7 & 405$\pm$29 & 52.1 \\
GA & 268$\pm$13 & 42$\pm$9 & 255$\pm$13 & 38$\pm$6 & 241$\pm$11 & 35$\pm$7 & 248$\pm$12 & 74.1 & 289$\pm$16 & 67.3 & 352$\pm$23 & 58.6 \\
PPO & 259$\pm$15 & 35$\pm$8 & 248$\pm$12 & 32$\pm$9 & 234$\pm$10 & 28$\pm$6 & 241$\pm$14 & 75.3 & 281$\pm$16 & 68.9 & 345$\pm$23 & 59.2 \\
QMIX & 251$\pm$15 & 31$\pm$6 & 242$\pm$10 & 29$\pm$7 & 229$\pm$10 & 25$\pm$5 & 235$\pm$10 & 76.8 & 274$\pm$15 & 70.2 & 336$\pm$21 & 60.8 \\
MAPPO & 243$\pm$12 & 28$\pm$7 & 235$\pm$11 & 26$\pm$5 & 223$\pm$9 & 22$\pm$5 & 228$\pm$11 & 78.4 & 265$\pm$12 & 72.1 & 324$\pm$19 & 62.5 \\
MAPPO+ER & 238$\pm$10 & 25$\pm$5 & 231$\pm$10 & 24$\pm$4 & 220$\pm$7 & 20$\pm$4 & 224$\pm$10 & 79.1 & 259$\pm$13 & 73.5 & 316$\pm$18 & 63.7 \\
GNN-Disp.\cite{c19_song2023} & 245$\pm$12 & 27$\pm$7 & 237$\pm$10 & 25$\pm$6 & 221$\pm$8 & 21$\pm$5 & 230$\pm$11 & 77.8 & 267$\pm$14 & 71.5 & 327$\pm$18 & 61.9 \\
MAPPO+Epi\cite{c17_pritzel2017} & 233$\pm$11 & 22$\pm$4 & 227$\pm$9 & 21$\pm$4 & 217$\pm$8 & 18$\pm$4 & 220$\pm$9 & 80.3 & 253$\pm$10 & 74.8 & 309$\pm$17 & 65.1 \\
\midrule
\textbf{GSEM (Ours)} & \textbf{218$\pm$9} & \textbf{14$\pm$4} & \textbf{214$\pm$8} & \textbf{13$\pm$3} & \textbf{208$\pm$7} & \textbf{12$\pm$3} & \textbf{209$\pm$10} & \textbf{83.6} & \textbf{237$\pm$9} & \textbf{78.4} & \textbf{278$\pm$15} & \textbf{71.2} \\
% \midrule
% Impr.\% & 6.4 & 36.4 & 5.7 & 38.1 & 4.1 & 33.3 & 5.0 & 4.1 & 6.3 & 4.8 & 10.0 & 9.4 \\
\bottomrule
\end{tabular}
}
\end{table*}

\subsection{Main Results}

Table~\ref{tab:main} summarizes the results across all scenarios. Two baseline comparisons are particularly informative. First, GNN-Dispatch achieves performance comparable to MAPPO despite using a graph neural network encoder, confirming that GNN-based encoding of the \textit{current} state alone does not improve adaptation to dynamic disturbances. Second, MAPPO+Epi benefits from episodic retrieval and outperforms MAPPO+ER, yet its reliance on flat state vectors limits its ability to preserve the relational recovery structures within coordination episodes. GSEM closes this gap by storing recovery motifs as graphs, consistently achieving the best performance with makespan reductions of $4.1$\%--$6.4$\% and adaptation speed improvements of $33$\%--$38$\% relative to MAPPO+Epi across all single-disturbance scenarios.

\begin{figure}[t]
    \centering
    \includegraphics[width=0.95\columnwidth]{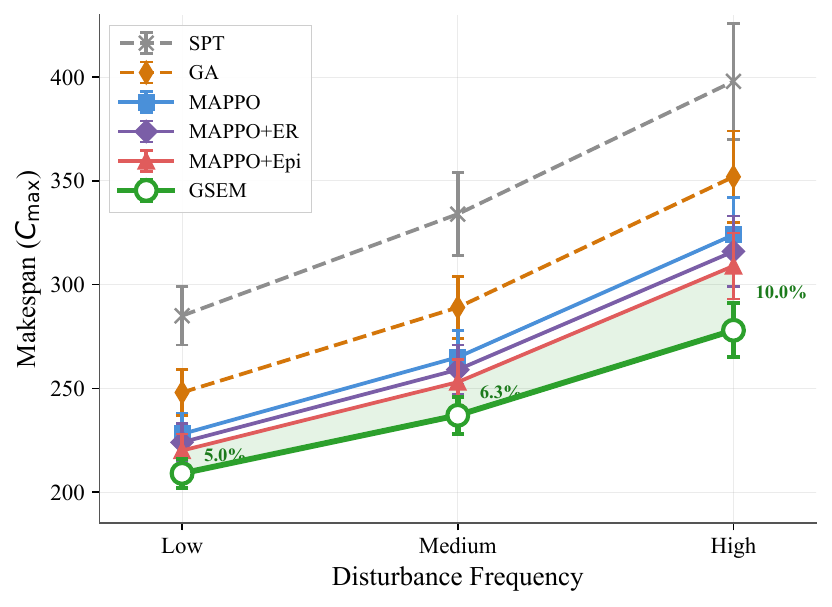}
    \caption{Makespan under increasing disturbance frequency (mixed disturbances).}
    \label{fig:freq}
\end{figure}

The advantage of GSEM becomes more pronounced under mixed disturbances (Fig.~\ref{fig:freq}). As disturbance frequency increases from Low to High, the makespan improvement over MAPPO+Epi grows from $5.0$\% to $10.0$\%. Under high-frequency conditions, new disturbances arrive before previous ones are fully resolved, creating cascading disruptions. GSEM mitigates this by retrieving relevant recovery patterns from its memory pool.

\begin{figure}[t]
    \centering
    \includegraphics[width=0.95\columnwidth]{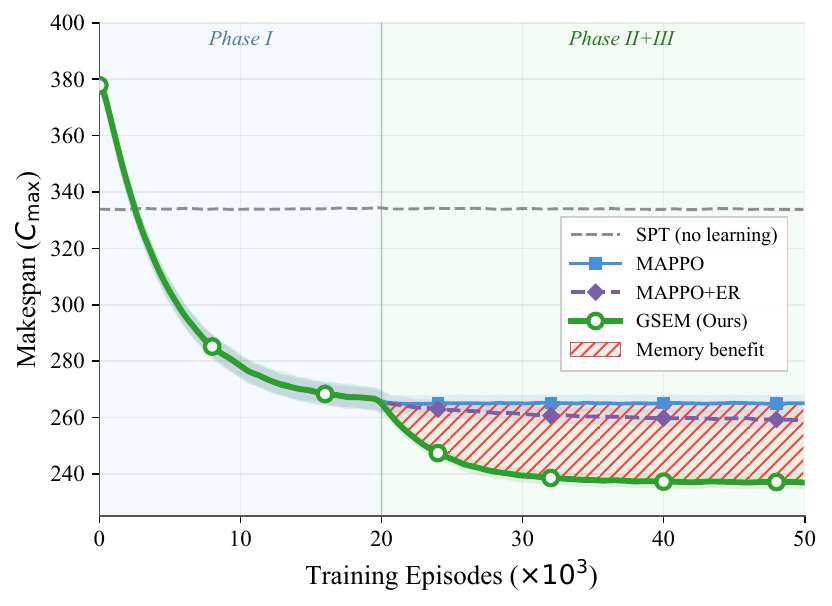}
    \caption{Training curves under mixed disturbances (medium frequency).}
    \label{fig:curve}
\end{figure}

\subsection{Training Convergence}

Fig.~\ref{fig:curve} presents the learning curves under mixed disturbances at medium frequency. In Phase~I, all three methods exhibit similar convergence. Once Phase~II+III begins, MAPPO and MAPPO+ER plateau, while GSEM continues to improve as the memory pool accumulates high-quality coordination episodes. This confirms that reusing past recovery structures provides a complementary learning signal that reactive policy optimization alone cannot capture.

\subsection{Ablation Study}

Table~\ref{tab:ablation} reports ablation results under mixed disturbances at medium frequency. The results directly validate the necessity of graph-structured recovery experience reuse. First, replacing graph memory with flat vectors (w/o Graph) increases makespan by $6.3$\%, confirming that the benefit comes from preserving relational recovery structure, not from memory augmentation alone. Second, replacing similarity-based retrieval with random retrieval (Random Retrieval) causes the largest degradation ($8.0$\%), demonstrating that the system genuinely reuses past recovery motifs rather than benefiting from generic experience augmentation, as irrelevant experiences actively mislead agents. The gating mechanism and memory-conditioned communication contribute $4.2$\% and $5.1$\% improvements, respectively, by preventing negative transfer and enabling system-level experience sharing.

\begin{table}[t]
\caption{Ablation Study (10$\times$10, Mixed, Med. Freq.). $T$: mean tardiness.}
\label{tab:ablation}
\centering
\resizebox{\columnwidth}{!}{
\begin{tabular}{lcccc}
\toprule
Variant & $C_{\max}$ & Adapt. & $U$(\%) & $T$ \\
\midrule
\textbf{GSEM (Full)} & \textbf{237$\pm$9} & \textbf{14$\pm$3} & \textbf{78.4} & \textbf{42$\pm$8} \\
\midrule
w/o Graph (flat vector) & 252$\pm$10 & 20$\pm$5 & 74.1 & 56$\pm$13 \\
Random Retrieval & 256$\pm$14 & 22$\pm$6 & 73.2 & 61$\pm$11 \\
w/o Decay (FIFO evict) & 244$\pm$11 & 17$\pm$3 & 76.5 & 49$\pm$10 \\
w/o Gate (concat) & 247$\pm$9 & 18$\pm$5 & 75.8 & 51$\pm$11 \\
w/o Communication & 249$\pm$12 & 19$\pm$4 & 75.2 & 53$\pm$9 \\
w/o Contrastive & 242$\pm$9 & 16$\pm$4 & 77.0 & 47$\pm$10 \\
\bottomrule
\end{tabular}
}
\end{table}

\begin{table}[t]
\caption{Cross-Disturbance Transfer (Makespan, 10$\times$10, Med. Freq.)}
\label{tab:transfer}
\centering
\resizebox{\columnwidth}{!}{
\begin{tabular}{lcccc}
\toprule
\multirow{2}{*}{Memory Source} & \multicolumn{3}{c}{Test Disturbance} & \multirow{2}{*}{Avg.} \\
\cmidrule(lr){2-4}
 & Breakdown & Urgent & Time Var. & \\
\midrule
No memory (MAPPO) & 243$\pm$12 & 235$\pm$11 & 223$\pm$9 & 233.7 \\
\midrule
Breakdown & \textit{218$\pm$9} & 228$\pm$10 & 216$\pm$7 & 220.7 \\
Urgent Job & 232$\pm$11 & \textit{214$\pm$8} & 215$\pm$7 & 220.3 \\
Time Variation & 235$\pm$10 & 230$\pm$9 & \textit{208$\pm$7} & 224.3 \\
\midrule
Mixed (all types) & \textbf{216$\pm$10} & \textbf{212$\pm$8} & \textbf{206$\pm$7} & \textbf{211.3} \\
\bottomrule
\end{tabular}
}
\end{table}

\subsection{Computational Overhead}

Compared to MAPPO, GSEM increases per-step inference time by approximately $18$\% ($3.4$\,ms to $4.0$\,ms per agent) and total training time by $31$\% ($14.2$\,h vs.\ $10.8$\,h for $50$K episodes). The memory pool with $|\mathcal{P}|_{\max}{=}500$ requires approximately $120$\,MB of GPU memory. These overheads are modest and practical for manufacturing control where decision intervals are on the order of seconds. To verify scalability, we tested on a $15{\times}15$ instance with $8$ agents: GSEM achieves $365{\pm}14$ vs.\ $394{\pm}16$ for MAPPO+Epi ($7.4$\% improvement), confirming the advantage is maintained at larger scale.

\subsection{Cross-Disturbance Transferability}

Table~\ref{tab:transfer} reports transfer results where the memory pool is populated from a single disturbance type and evaluated on different types without retraining.

Even under mismatched source-target conditions, GSEM consistently outperforms memoryless MAPPO, indicating that graph-structured memory captures transferable recovery patterns that generalize across disturbance types. The mixed-source memory achieves the lowest average makespan ($211.3$ vs.\ $233.7$ for MAPPO), confirming that diverse disturbance exposure yields a more versatile memory pool.

%%%%%%%%%%%%%%%%%%%%%%%%%%%%%%%%%%%%%%%%%%%%%%%%%%%%%%%%%%%%%%%%%%%%%%%%%%%%%%%%
\section{CONCLUSION}

This paper presented GSEM, a graph-structured experiential memory framework for multi-agent coordination in dynamic manufacturing. By storing past coordination episodes as heterogeneous relational graphs and retrieving structurally similar experiences, GSEM enables experience-guided recovery rather than learning from scratch.

Experiments on DFJSP benchmarks demonstrated makespan reductions of $4.1$\%--$10.0$\% and adaptation time improvements of $33$\%--$38$\% over the strongest baselines, with the advantage being most pronounced under high-frequency mixed disturbances. Ablation studies confirmed that graph-structured encoding and similarity-based retrieval are the two most critical components. Cross-disturbance transfer experiments further showed that the learned recovery patterns generalize across disturbance types. Future directions include scaling to larger instances, integrating human operator feedback, and extending the framework to other dynamic combinatorial optimization domains.

\addtolength{\textheight}{-12cm}

%%%%%%%%%%%%%%%%%%%%%%%%%%%%%%%%%%%%%%%%%%%%%%%%%%%%%%%%%%%%%%%%%%%%%%%%%%%%%%%%
\section*{ACKNOWLEDGMENT}
This work was supported by Chuchiang Data Co., Ltd., Shanghai, China. The authors gratefully acknowledge the computational resources provided by Chuchiang Data.

%%%%%%%%%%%%%%%%%%%%%%%%%%%%%%%%%%%%%%%%%%%%%%%%%%%%%%%%%%%%%%%%%%%%%%%%%%%%%%%%

\end{document}